\documentclass[]{spie}

\usepackage{amsmath,amsfonts,amssymb}
\usepackage{graphicx}
\usepackage[colorlinks=true, allcolors=blue]{hyperref}
\usepackage{subcaption}
\usepackage{textcomp}

\graphicspath{{res/}}

\title{Exploiting Clinically Available Delineations for CNN-based Segmentation in Radiotherapy Treatment Planning}

\author[a,b,d]{Louis D. van Harten}
\author[a,d]{Jelmer M. Wolterink}
\author[b]{Joost J.C. Verhoeff}
\author[a,c,d]{Ivana I\v{s}gum}

\affil[a]{Department of Biomedical Engineering and Physics, Amsterdam University Medical Center, University of Amsterdam, The Netherlands}
\affil[b]{Department of Radiotherapy, University Medical Center Utrecht, Utrecht, The Netherlands}
\affil[c]{Department of Radiology and Nuclear Medicine, Amsterdam University Medical Center, University of Amsterdam, The Netherlands}
\affil[d]{Image Sciences Institute, University Medical Center Utrecht, Utrecht, The Netherlands}

\authorinfo{Further author information: (Please send correspondence to L.D. van Harten)\\E-mail: l.d.vanharten@amsterdamumc.nl, Telephone: +31 20 56 65206}

\begin{document}

\maketitle

\begin{abstract} 
    Convolutional neural networks (CNNs) have been widely and successfully used for medical image segmentation. However, CNNs are typically considered to require large numbers of dedicated expert-segmented training volumes, which may be limiting in practice. This work investigates whether clinically obtained segmentations which are readily available in picture archiving and communication systems (PACS) could provide a possible source of data to train a CNN for segmentation of organs-at-risk (OARs) in radiotherapy treatment planning. In such data, delineations of structures deemed irrelevant to the target clinical use may be lacking. To overcome this issue, we use multi-label instead of multi-class segmentation. We empirically assess how many clinical delineations would be sufficient to train a CNN for the segmentation of OARs and find that increasing the training set size beyond a limited number of images leads to sharply diminishing returns. Moreover, we find that by using multi-label segmentation, missing structures in the reference standard do not have a negative effect on overall segmentation accuracy. 
    These results indicate that segmentations obtained in a clinical workflow can be used to train an accurate OAR segmentation model.
\end{abstract}


\section{Introduction}
Recent years have seen substantial progress in automatic medical image segmentation. These advances can be primarily attributed to the emergence of convolutional neural networks (CNNs) \cite{ciresan2012deep,Litj17}.
CNN-based segmentation is typically performed in a fully supervised setting, where a network is trained based on available segmentation maps in a training set. In such settings, the performance varies based on the network architecture and hyper-parameters, the optimization procedure during training, and on the size and quality of the training set.
A number of recent benchmarks have shown that most state-of-the-art CNN-based methods achieve comparable results when applied to the same medical image data set. For example, a recent challenge in cardiac cine MR images found that for left ventricle segmentation, there were no significant differences between the top 8 methods, even though all used different architectures, hyper parameters, and optimization schemes.\cite{Bern18}. All methods used the same training data, suggesting that the properties of the available training data used to train a convolutional neural network may be among the most important factors for the network performance. 

For medical image segmentation, creating a labeled training set typically entails time-consuming and costly annotation by medical experts. As a consequence, the number of available labeled examples in medical image training sets is generally much lower than the number of labeled examples in training sets for natural images. This issue is exacerbated by the large variety of medical imaging modalities and sequences, which generally means a completely new data set is required for every medical segmentation problem. 
A possible solution is to use data that is produced and manually labeled as part of a clinical workflow. For example, in radiotherapy, manual segmentations of organs-at-risk (OARs) are routinely made for treatment planning. In this work, we investigate whether -- instead of obtaining a data set of dedicated segmentations by a clinical expert -- these readily available clinical segmentations could be used to train a CNN for automatic segmentation of OARs. One challenge to overcome is that this data often lacks delineations of structures deemed irrelevant for the clinical task. For example, organs that are far away from a tumor, and hence not at risk of irradiation, are often not segmented. 

The use of partially segmented training volumes raises interesting methodological questions, as there is no unambiguous definition of ``background'' in such volumes. This problem has previously been addressed using conventional machine learning techniques~\cite{moeskops_evaluation_2015}.
Recently, CNN training strategies have been adapted for training with missing annotations by considering segmentation to be a \textit{multi-label} instead of a \textit{multi-class} problem~\cite{petit_handling_2018}. While multi-class segmentation requires all voxels to exclusively belong to the background class or to one of the foreground classes, a multi-label segmentation model produces a result for each of the foreground classes independently from the others. This property can be exploited to train the network with images for which not all classes have a ground truth label.

Here, we perform an empirical study in which we systematically investigate the number of reference delineations that is necessary to achieve adequate model performance for OAR segmentation, and whether similarly adequate results can be obtained using reference delineations with missing structures. By training and evaluating the performance of 96 CNNs trained on different subsets of our data set, we assess the feasibility of developing a successful OAR segmentation model using varying amounts of clinically available segmentations with varying levels of completeness.

\begin{figure}[t]
    \centering
    \vspace{0.35cm}
    \includegraphics[width=0.99\textwidth, clip, trim=0 0mm 0 0]{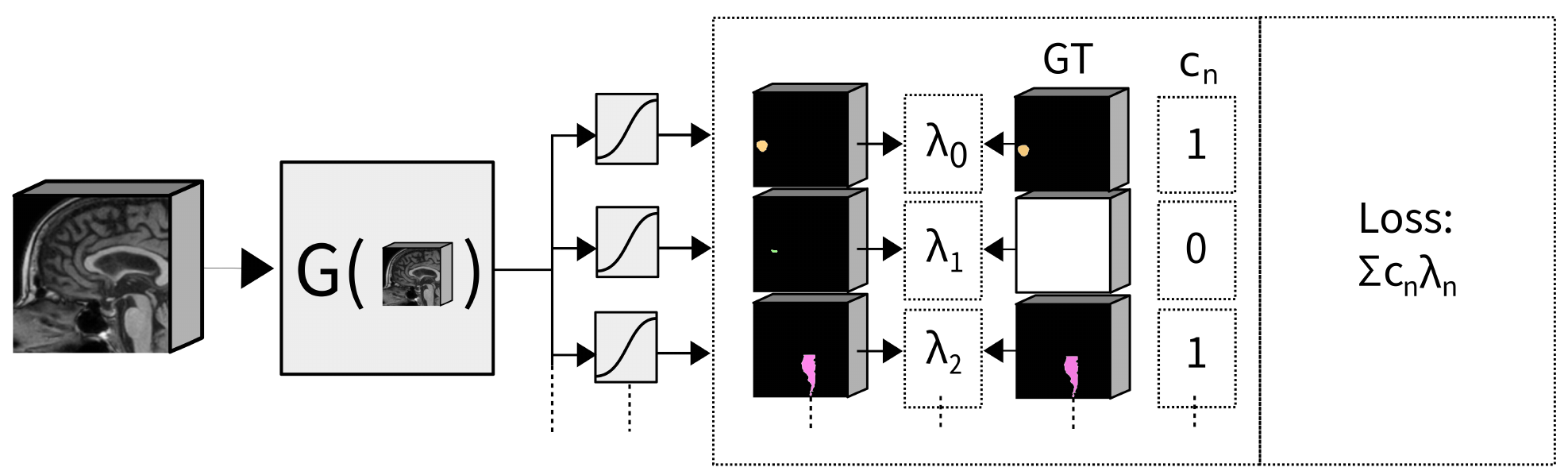} 
    \vspace{0.35cm}
    \caption{A schematic overview of the multi-label training process. Function $G$ represents a convolutional neural segmentation network with $N$ outputs, corresponding to each class. For each class $n$, a sigmoid function generates a probability map that is compared to a binary reference image through binary cross-entropy, resulting in a loss $\lambda_n$. The total loss is the sum over all $\lambda_n$, weighted by $c_n\in\{0,1\}$ depending on the presence of class $n$ in the ground truth (GT) of the training sample.}\label{fig:pipeline_architecture}
    \vspace{0.35cm}
\end{figure}

\section{Data}\label{sec:data}
With permission of the local medical ethics board, we included brain MRI studies of 52 patients undergoing radiotherapy treatment planning. All patients received a T\textsubscript{1}-weighted MR scan at the University Medical Center Utrecht (Utrecht, the Netherlands). Volumes were acquired using a Philips Ingenia 1.5T MR system with a voxel size of $1.1\times1.1\times1.0$ mm$^3$, 8\textdegree~ flip angle, 7 ms repetition time, and 3.1 ms echo time. Scans were reconstructed to a voxel size of $0.9\times0.9\times1.0$ mm$^3$. Patients were scanned in an immobilising mask, ensuring a similar orientation of the head for all patients.

This data set includes annotations of the brain stem, pituitary gland and optic chiasm, and the left and right optic nerves, eyes, cochleas, and lenses acquired as part of RT treatment planning. OARs were typically segmented only if they were in clinically relevant proximity to the clinical target volume for the RT treatment. On average 9.4$\pm$1.8 out of 11 possible OARs were annotated in each patient. For 15 out of 52 patients, all OARs were available. Delineations were made on CT images and propagated to corresponding MR images.
This regularly led to over- and under-segmentation when visualised in the MR image. Given that these are representative of clinical segmentations, we used these potentially suboptimal segmentations for training. However, as the results must be evaluated on a ground truth, such errors would interfere with the evaluation. Hence, a clinical expert corrected all manual segmentations in a subset of 20 volumes, which was used as a test set.

\begin{figure}[t]
    \centering
    \includegraphics[width=0.70\textwidth, clip, trim=0 0mm 38mm 0]{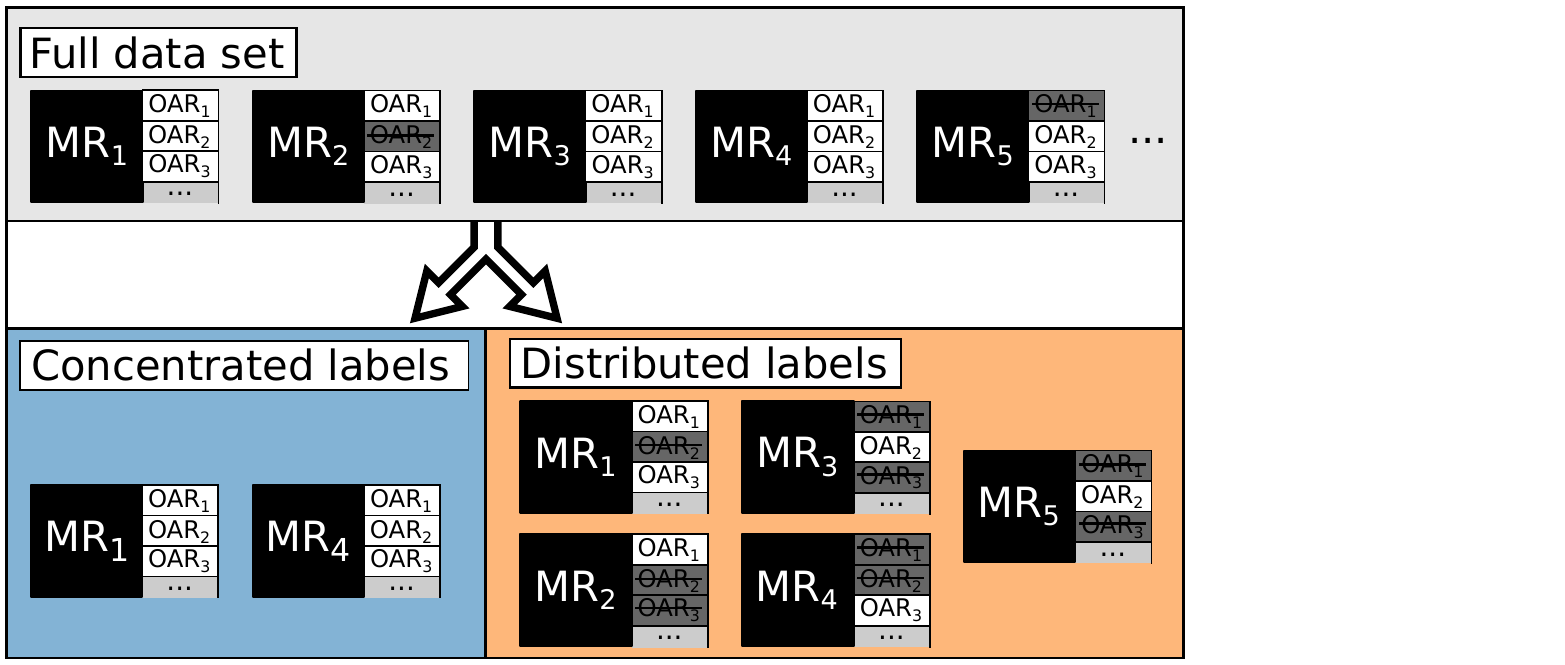} 
    \vspace{0.35cm}
    \caption{An illustration of the two sampling methods used in this work, visualised for subsets of size $M$=2. The full data set contains some MR images for which not all class labels are available (indicated by strike-through). \textit{Concentrated label} subsets contain $M$ volumes for which all labels are available. \textit{Distributed label} subsets contain more MR volumes, but labels for each class are only available in $M$ volumes.}\label{fig:sampling_diagram}
\end{figure}

\begin{figure}[b]
    \centering
    \includegraphics[width=0.99\textwidth, clip, trim=0 0mm 0 0]{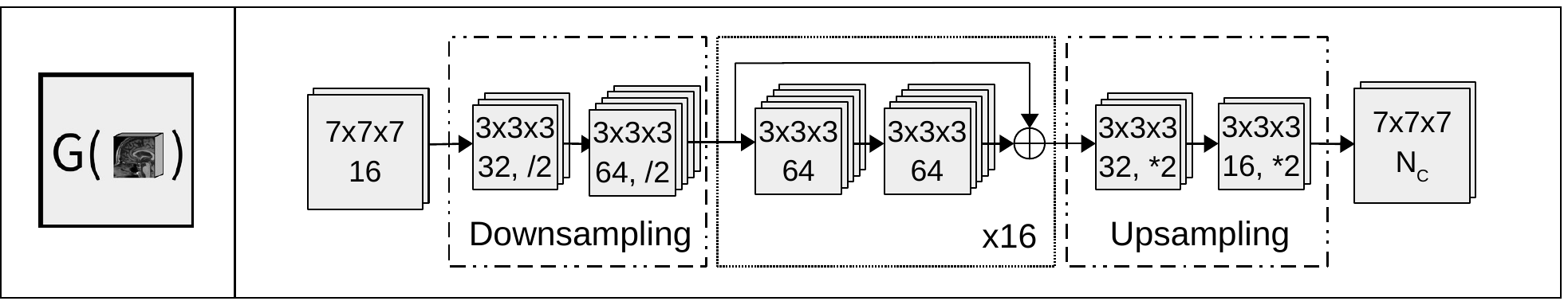} 
    \vspace{0.35cm}
    \caption{A detailed view of the 3D CNN architecture used in this work, indicated as function $G$ in Figure~\ref{fig:pipeline_architecture}. Each unit in this figure represents a 3D convolution layer with a ReLu activation, with numbers in each unit indicating the kernel size and number of filter maps. The down- and upsampling paths are implemented using strided convolutions (indicated by `/2') and transposed convolutions (indicated by `*2') respectively. $N_c$ indicates the amount of classes and is equal to 11 in this work.}\label{fig:detailed_architecture}
\end{figure}

\section{Methods}\label{sec:methods}
We perform a series of experiments to address two separate but related research questions.
We investigate whether incomplete segmentations obtained from a clinical workflow are an appropriate substitute for a dedicated training set for training a segmentation network for various brain structures. Additionally, we investigate the impact of the number of segmented training volumes on the performance of such a network, and whether the required number of segmented volumes increases when only part of the target structures are segmented in each training image.

To address these questions, we perform 96 experiments in which we train CNNs with sampled subsets of the available training data. A subset of size $M$ is defined as a data set in which each structure has been segmented $M$ times. Subsets are randomly sampled in two distinct ways, as illustrated in Fig.~\ref{fig:sampling_diagram}. In the \textit{concentrated labels} setting, we sample $M$ volumes with full reference segmentations, which is equivalent to selecting the MR volumes of $M$ patients. In the \textit{distributed labels} setting, a subset includes all training volumes, but for each class, labels are only included in $M$ randomly selected volumes.
In this setting, a training subset contains more volumes of different patients, but labels for only part of the structures in each volume are available. Subsets in the distributed labels setting were pseudo-randomly sampled to evenly spread the labels over as many volumes as possible.  In both settings, the trained networks see the same number of labeled structures and -- assuming similarly sized structures among patients -- a similar number of training voxels. This equates to an approximately equal amount of work required by a clinical expert to create the training sets, which allows us to compare the results fairly. As smaller subsets can be sampled in many ways, we repeat experiments multiple times with different subsets of the same size.

In all experiments, we use the same 3D fully convolutional network with residual connections, adapted from an existing 2D network\cite{he_deep_2015}. This network was recently shown to exhibit competitive performance in a challenge on OAR segmentation in thoracic CT images\cite{vanharten2019automatic}. Its architecture is shown in Fig.~\ref{fig:detailed_architecture}. The network contains two strided convolutional downsampling layers, followed by 16 residual blocks and two transposed convolutional upsampling layers. The residual blocks are implemented using the updated residual configuration proposed in He et al.~\cite{he2016identity}.
Instead of a softmax activation function, which is typically used in multi-class segmentation problems, the output layer contains one sigmoid activation function per class (as shown in Fig.~\ref{fig:pipeline_architecture}).
By using a sigmoid instead of a softmax output activation function, any $N$-class segmentation problem can be modelled as a combination of $N$ binary label segmentation problems\cite{petit_handling_2018}.
A class loss $\lambda_n$ can be calculated for each binary label separately; the total loss is calculated as the sum over all class losses, weighted by an optional weighting factor $c_n$. If during training class $n$ is not present in the reference segmentation of the current training volume, $c_n$ is set to 0. This amounts to ignoring the loss component corresponding to this class for the current training volume. In this work, $c_n$ is otherwise set to 1 for all classes.

\begin{figure}[t]
    \centering
    \newcommand{\detgraphheight}{7.5cm}
    \includegraphics[height=\detgraphheight, clip, trim=0 2mm 0 0]{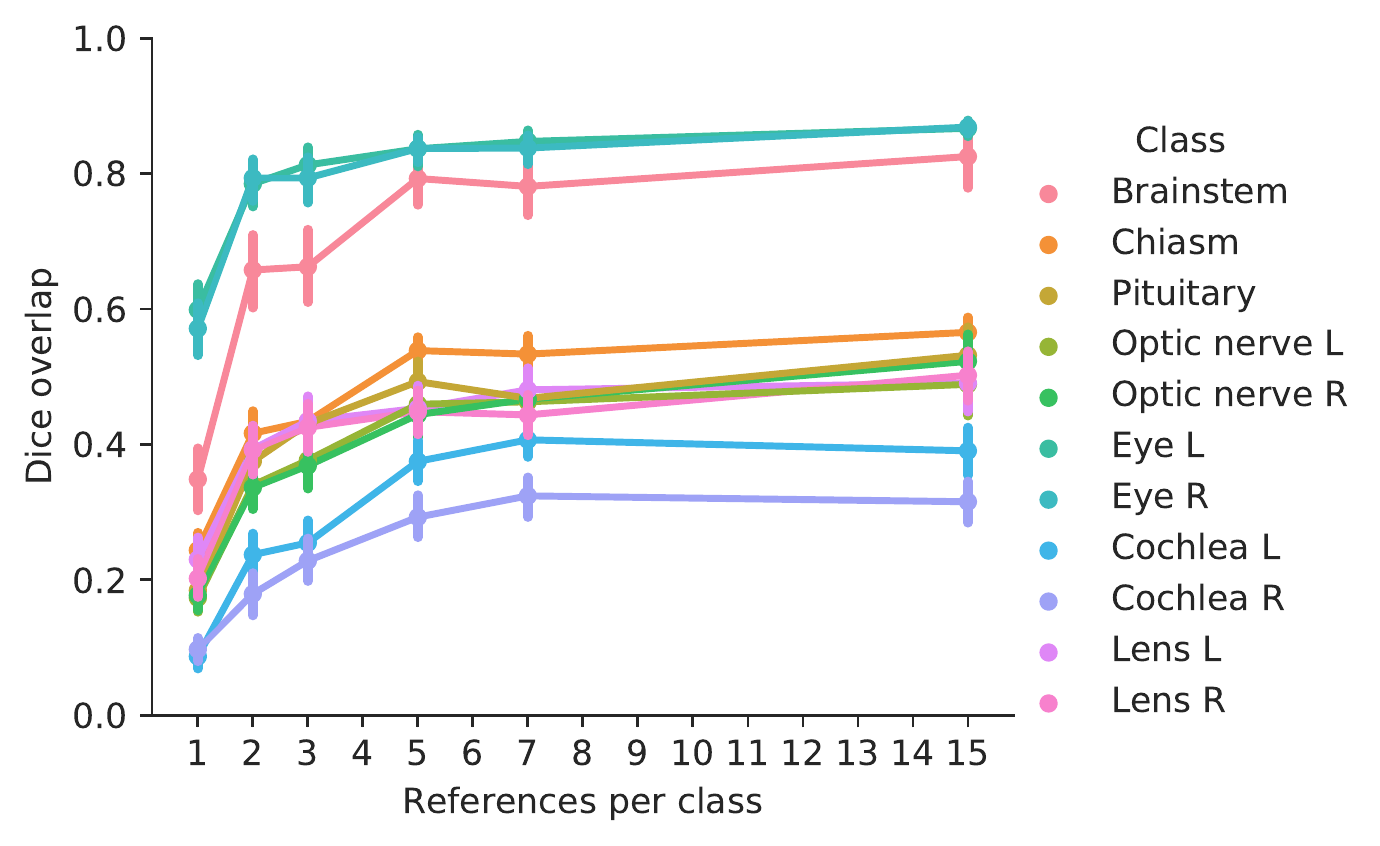} 
    \vspace{0.15cm}
    \caption{Network performance in terms of Dice similarity coefficient as a function of the number of delineated ground truth references per class, when trained in the concentrated (i.e.: fully segmented) training configuration. Data points on the graph are the average Dice score for each OAR on the test set, averaged over the results from each network trained on the same training subset size. Vertical bars indicate the 95\% confidence interval.
    }\label{fig:results_mrct_concentrated}
\end{figure}

\section{Experiments and Results}
From the full data set, 30 volumes were used for training, 2 volumes were used for validation, and the remaining 20 volumes were used for testing.
We trained 96 networks on the OAR set: 48 on concentrated subsets and 48 on distributed subsets, as described in Sec.~\ref{sec:methods}. In the training set, 15 volumes included reference delineations for all 11 target OARs and could be used to sample concentrated subsets. No data augmentation was used in any of the experiments. All networks were trained with the Adam optimizer (learning rate: 0.001) for 15000 iterations with batches of four cubic 64\textsuperscript{3}-voxel patches per iteration. Training was done on a shared computing cluster containing various consumer-grade NVIDIA GPUs; training times ranged between 4 and 12 hours per network, depending on the load on the cluster.

\begin{figure*}[t]
    \centering
    \newcommand{\stripheight}{6.75cm}
        \begin{subfigure}[t]{0.48\textwidth}
        \includegraphics[height=\stripheight, clip, trim=0 0mm 0 0]{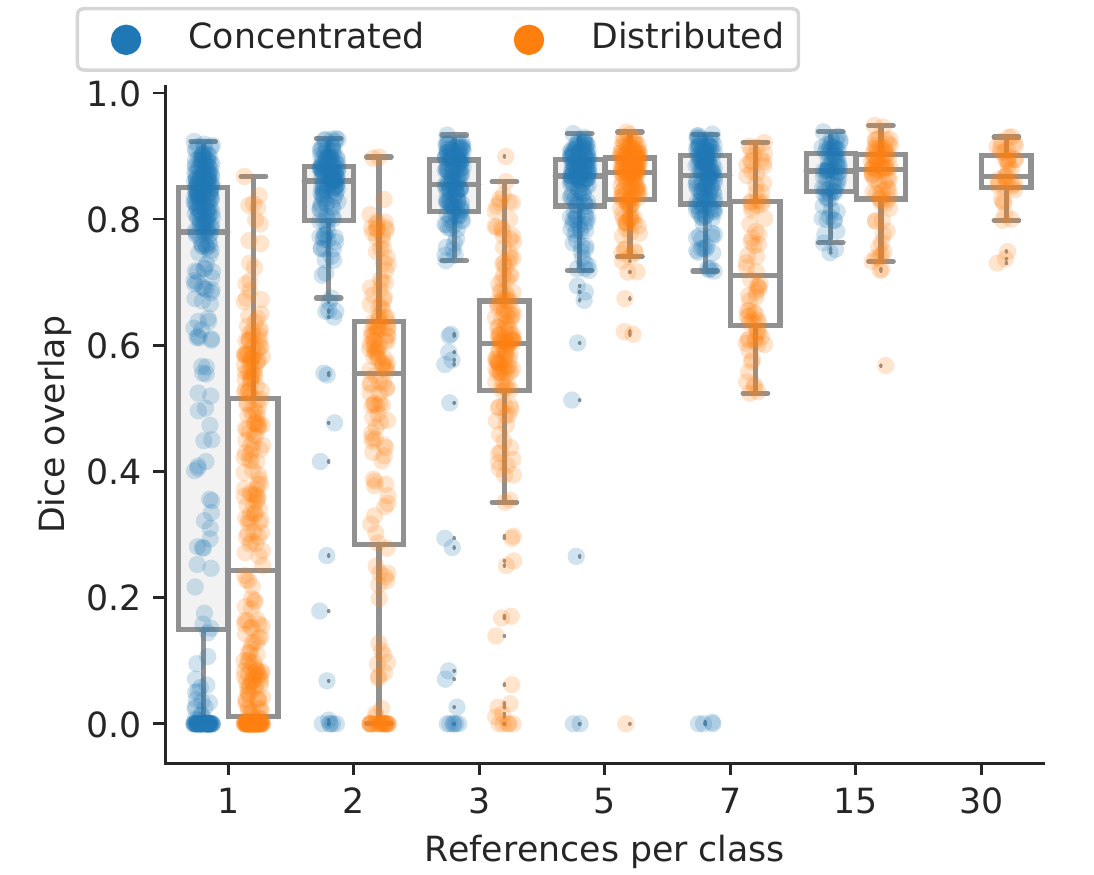}
        \caption{Right eye} \label{fig:results_mrct_reye}
    \end{subfigure}
    ~
    \begin{subfigure}[t]{0.48\textwidth}
        \includegraphics[height=\stripheight, clip, trim=0 0mm 0 0]{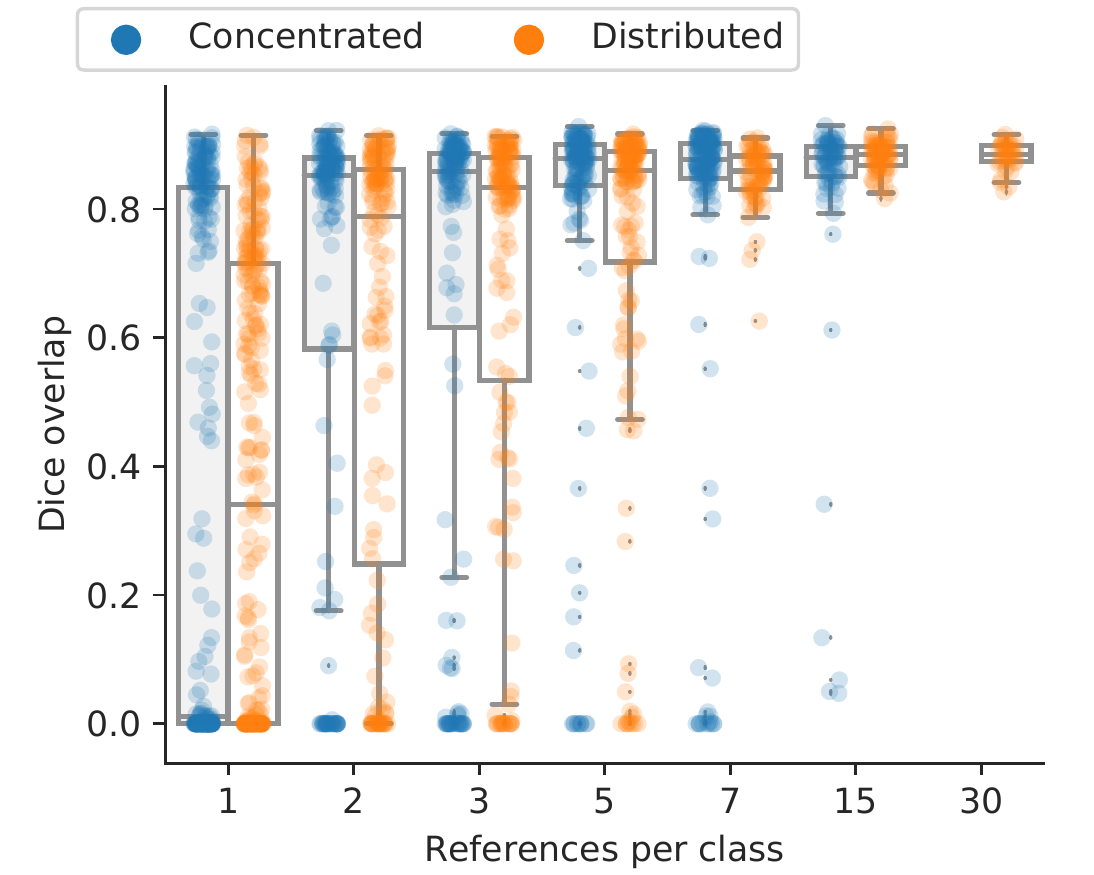}
        \caption{Brain stem} \label{fig:results_mrct_brainstem}
    \end{subfigure}
    ~
    \vspace{0.25cm}

    \caption{Network performance in terms of Dice similarity coefficient as a function of the number of delineated ground truth references per class, for the right eye (a) and for the brain stem (b). Rightmost results in both figures signify all available reference segmentations were used ($\leq$30 references per class). The results from all experiment repetitions are plotted together. Note that the $x$-axis is not linear, but that the number of references approximately doubles in each step.} \label{fig:results_all_concentrated}
\end{figure*}

Fig.~\ref{fig:results_mrct_concentrated} shows the average Dice similarity coefficients attained by networks trained on concentrated (i.e. fully segmented) training sets. The results show that, as may be expected, the performance increases when more training volumes are used. The improvement of the average performance is most pronounced up until five reference segmentations per structure. The results for most classes still slightly improve when increasing the training set size further, albeit with sharply diminishing returns. 

We assessed whether the same number of reference segmentations can also be provided to the CNN in a distributed fashion, i.e. using volumes in which only a subset of structures has been segmented. We compare the results for the concentrated and distributed settings for the right eye and the brain stem in Fig.~\ref{fig:results_all_concentrated}. These figures show that for the larger training sets, the performance of networks trained in both settings is similar, although networks trained in the distributed setting produce a smaller number of worst-case outliers. 

Interestingly, the results show that the networks trained on small sets of distributed data perform substantially worse on the eye, whereas they are mostly comparable with the networks trained on concentrated data for the brain stem. This discrepancy could be explained by the presence of visual ambiguity in the classes that have contralateral equivalents. Because of our pseudo-random sampling in the distributed experiment setting, it was highly unlikely that reference segmentations for both versions of a symmetric OAR would be included for the same patient. As a result, these networks perform worse at distinguishing symmetric OARs from their contralateral equivalents. Intuitively, a right eye is difficult to distinguish from a left eye based on local geometry: the networks may fail to learn the distinction unless labels for a matching set of eyes is available.

\section{Discussion}
In this study, we aimed to answer two research questions. First, we evaluated the number of segmented volumes required to train an adequate OAR segmentation network using clinically obtained data. We observed a saturation point between five and seven labeled examples in the training sets (depending on the class) after which an increased data set size showed sharply diminishing returns on the network performance. This challenges the common assumptions concerning the large amounts of data required for training a deep neural network. Although these results are promising, they were acquired using a relatively small test set in a single segmentation task; further research is needed to investigate the extent to which our findings apply to different segmentation tasks.

Second, we evaluated whether a similarly performing network could be trained using a data set in which only some classes were segmented in each volume, as is generally the case with clinically performed segmentations. In this setting, we observed a large discrepancy in performance on the symmetric OARs with contralateral equivalents in networks trained on small training sets. Our results imply that the networks have trouble learning to discriminate between visually similar structures unless both are segmented in the same training volume.

It should be noted that the problem illustrated above is only present when training on subsets where the majority of segmented volumes only contain one version of the symmetric OARs. In the full data set, segmentations of contralaterally equivalent organs are usually both present if a segmentation for at least one of them is available. This discrepancy could be considered an artifact of our sampling method. However, the presence of unsegmented visually ambiguous structures is not unthinkable in other clinically obtained data sets. For example, similar problems could emerge with partially segmented vertebral columns, or abdominal images where only part of the large intestine is delineated. Future work could investigate the merit of cropping such unsegmented visual ambiguities out of the training images before training the network.

\section{Conclusion}
In this work, we have addressed the common assumption that segmentation CNNs require large amounts of training data and we have investigated whether routinely acquired clinical segmentations can be used to train an OAR segmentation model instead of a dedicated training set. 
We found that training such networks on a small number of incomplete clinical segmentations is feasible, as long as there are no clear ambiguities between classes. We have shown that this limitation can be overcome by increasing the size of the training set.

\section{New or breakthrough work to be presented}
We have shown that it is possible to train an accurate OAR segmentation network with a small training set of clinically acquired delineations, without any data augmentation. Our results show that as long as there is little ambiguity in the class definitions, it is possible to train such a network even if part of the target class delineations is missing in each of the training volumes.

\bibliography{spie_seg}
\bibliographystyle{spiebib}

\end{document}